\documentclass[final]{cvpr}

\usepackage{times}
\usepackage{epsfig}
\usepackage{graphicx}
\usepackage{amsmath}
\usepackage{amssymb}
\usepackage{multirow}
\usepackage{enumitem}
\usepackage{tabulary}

\usepackage{xcolor, colortbl}
\usepackage{booktabs}

\usepackage[pagebackref,breaklinks,colorlinks]{hyperref}

\def\cvprPaperID{20103} %
\def\confName{CVPR}
\def\confYear{2026}
\graphicspath{{./figs/}{./figs/result/}}

\begin{document}

\title{MOSAIC-GS: Monocular Scene Reconstruction via Advanced Initialization for Complex Dynamic Environments}

\author{Svitlana Morkva$^{1}$
\hspace{7mm}
Maximum Wilder-Smith$^{1}$
\hspace{7mm}
Michael Oechsle$^{2}$
\hspace{7mm}
Alessio Tonioni$^{2}$
\\[1ex]
Marco Hutter$^{1}$
\hspace{7mm}
Vaishakh Patil$^{1}$
\\[1ex]
$^{1}$ETH Z\"urich
\hspace{7mm}
$^{2}$Google\\
{\tt\small {\{smorkva, mwilder, mahutter, patilv\}@ethz.ch}} \hspace{7mm} 
{\tt\small {\{michaeloechsle,alessiot\}@google.com}}
}

\twocolumn[{
\maketitle
\begin{center}
    \vspace{-1em}
    \includegraphics[width=\textwidth]{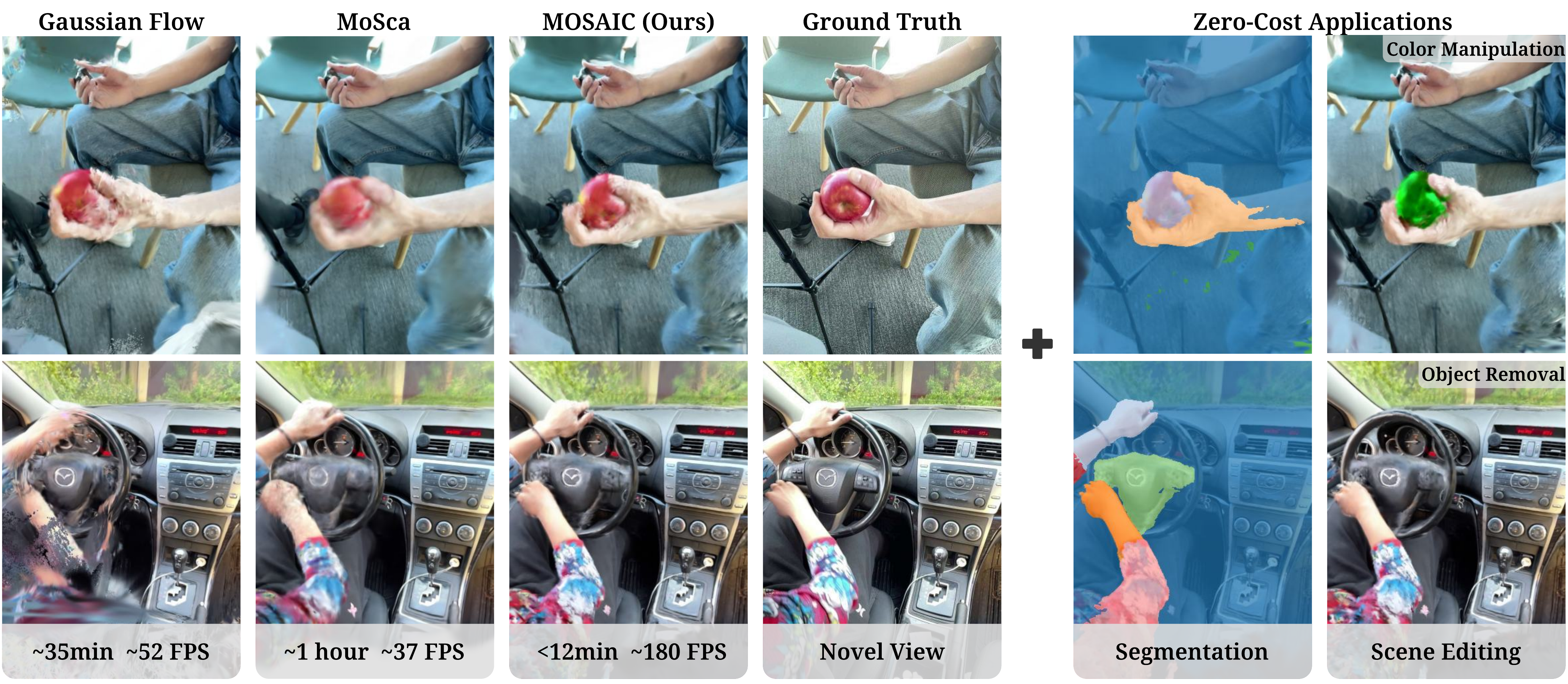}
    \captionof{figure}{
        MOSAIC-GS produces high-quality reconstructions of dynamic scenes, capturing fine details in dynamic regions more accurately while significantly reducing training and rendering time compared to prior methods. Beyond reconstruction, our approach enables dynamic scene segmentation and editing, allowing objects to be removed, isolated, or manipulated without any additional computational cost.
    }
    \label{fig:teaser}
\end{center}
\vspace{1em}
}]

\begin{abstract}
We present MOSAIC-GS, a novel, fully explicit, and computationally efficient approach for high-fidelity dynamic scene reconstruction from monocular videos using Gaussian Splatting.
Monocular reconstruction is inherently ill-posed due to the lack of sufficient multiview constraints, 
making accurate recovery of object geometry and temporal coherence particularly challenging. 
To address this, we leverage multiple geometric cues, such as depth, optical flow, dynamic object segmentation, and point tracking. 
Combined with rigidity-based motion constraints, these cues allow us to estimate preliminary 3D scene dynamics during an initialization stage.
Recovering scene dynamics prior to the photometric optimization reduces reliance on motion inference from visual appearance alone, which is often ambiguous in monocular settings.
To enable compact representations, fast training, and real-time rendering while supporting non-rigid deformations, the scene is decomposed into static and dynamic components. Each Gaussian in the dynamic part of the scene is assigned 
a trajectory represented as time-dependent Poly-Fourier curve for parameter-efficient motion encoding.
We demonstrate that MOSAIC-GS achieves substantially faster optimization and rendering compared to existing methods,
while maintaining reconstruction quality on par with state-of-the-art approaches across standard monocular dynamic scene benchmarks.
\end{abstract}

\section{Introduction}
Recent advances in Neural Radiance Fields (NeRFs)~\cite{mildenhall2020nerf} and 3D Gaussian Splatting~\cite{kerbl3Dgaussians}
have revolutionized 3D scene reconstruction, enabling highly accurate and photorealistic modeling of complex environments.
However, these techniques are limited to static scenes, restricting their applicability in domains that involve dynamic environments,
such as robotics, autonomous driving, or digital-twin generation for virtual and augmented reality (VR/AR) applications.

Several recent methods extend these ideas to dynamic scenes~\cite{9710336, park2021hypernerf, 9578753, 9878989, 10550869, 10656774},
but the problem remains largely unsolved. Existing approaches often require extensive training time, substantial memory and storage
resources, and suffer from slow rendering, while still exhibiting noticeable artifacts in regions with complex motion.
Many of these methods also prioritize visual realism over physical consistency, leading to distortions when observed from unseen viewpoints.
This issue is particularly pronounced in settings with restricted camera viewpoints or monocular video acquisition,
where the limited field of view further complicates physically consistent reconstruction.
Such scenarios are common in robotics and embedded devices, where cameras typically provide only narrow or stereo views rather than wide multi-view coverage.
Moreover, the limited computational and memory resources of these platforms make compact and efficient scene representations essential.

To address these challenges, we present MOSAIC-GS, an explicit and computationally efficient framework for dynamic scene reconstruction from monocular videos.
MOSAIC-GS leverages a 3D Gaussian Splatting representation~\cite{kerbl3Dgaussians} constructed from different viewpoints of a dynamic scene acquired over an interval of time. Each view is paired with a camera pose and a depth estimate, obtained either by dedicated sensors or learned models.
A key insight of our work is that inferring scene dynamics from purely visual data during photometric optimization
is both inefficient and unreliable. Instead, we emphasize the importance of high-quality initialization, which has been shown to strongly influence reconstruction fidelity~\cite{kerbl3Dgaussians} and convergence speed~\cite{meuleman2025onthefly}.

Our pre-processing pipeline first focuses on robust identification and tracking of dynamic regions, enabling decomposition of the scene into static and dynamic components. This separation improves parameter efficiency and simplifies the subsequent optimization.
We then estimate preliminary Gaussian motions from all available geometric and physical cues in the video, leveraging recent advances in optical flow~\cite{raft}, object segmentation~\cite{ravi2024sam2}, and
per-point tracking~\cite{10376999, doersch2024bootstap} to extract per-point trajectories and object associations.
We further incorporate rigidity constraints during this stage to enable refinement of these motions and efficient inference even in unobserved regions of the scene.

To compactly encode motion, we represent these trajectories using time-dependent Poly-Fourier curves, following Lin \etal~\cite{lin2024gaussian}, 
instead of the per-frame deformation models commonly used in prior monocular methods~\cite{stearns2024marbles, som2024, 11094310}. These trajectories are then used to initialize the dynamic Gaussians.
Combined with disentanglement into static and dynamic components, this representation enables efficient modeling, fast training, real-time rendering,
and robust handling of non-rigid deformations. The proposed MOSAIC-GS framework is particularly effective in monocular settings, where limited viewpoints make motion inference from visual cues alone unreliable.

The main contributions of our work can be summarized as follows:
\begin{itemize}[leftmargin=*, topsep=1pt, itemsep=0pt, parsep=0pt, partopsep=0pt]
    \item An advanced pre-processing framework for dynamic Gaussian Splatting that improves reconstruction quality and significantly accelerates training.
    \item A robust pipeline for dynamic object detection and tracking across monocular video sequences.
    \item A novel trajectory refinement method that extracts and optimizes 3D motion using per-instance rigidity constraints.
    \item State-of-the-art training efficiency and inference speed, achieved through scene disentanglement, compact deformation representation, and accurate motion initialization.
\end{itemize}

\section{Related Works}

\subsection{Moving object detection}
Detection of moving objects in video remains a challenging problem, particularly under changing illumination and camera motion. 
However, advances in deep learning, particularly convolutional neural networks (CNNs)~\cite{726791, NIPS2012_c399862d, 7780459}, recurrent neural networks (RNNs)~\cite{rnn}, 
and more recently transformer-based architectures~\cite{NIPS2017_3f5ee243, dosovitskiy2021an}, have significantly improved robustness and accuracy. 
Building on these developments, recent motion detection frameworks integrate instance segmentation~\cite{8237584, redmon2015unified} of common dynamic objects, 
dense optical flow~\cite{Sun2018PWC-Net, raft, huang2022flowformer}, and epipolar geometry constraints to reliably detect motion even in challenging scenarios such as moving cameras or complex illumination~\cite{li2020neural, liu2023robust, 9197349, 11094310, Huang_2025_CVPR}.

\subsection{Novel-View Synthesis of Dynamic Scenes}

Novel-view synthesis aims to generate realistic images of a scene from previously unseen viewpoints.
A major breakthrough came with the introduction of Neural Radiance Fields (NeRFs)~\cite{mildenhall2020nerf}, which enabled high-fidelity 3D scene representations and have since been extended to capture dynamic content~\cite{9578753, 9711476, park2021hypernerf, li2020neural}. Despite their impressive quality, NeRF-based methods are computationally expensive and slow to render. To address this, hybrid approaches combine implicit neural fields with explicit 3D structures such as voxels or planes~\cite{10204118, 10204912, shao2023tensor4d, 10049689, attal2023hyperreel}, which improve efficiency but still struggle with real-time photorealistic rendering. 

This limitation has motivated the development of explicit point-based models such as 3D Gaussian Splatting~\cite{kerbl3Dgaussians}, which enable fast and high-quality rendering. Building on this foundation, recent works have extended Gaussian Splatting to dynamic scenes by modeling temporal motion in two main ways. The first line of work applies per-frame deformations, storing Gaussian transformations for each frame, either for all Gaussians or a smaller skeletal subset~\cite{10550869, 10657650, stearns2024marbles, 11094310, som2024}. 
These methods estimate per-point trajectories either during optimization or via point tracking, but they are memory-intensive and do not scale well to long sequences. 
The second direction adopts continuous motion modeling, where Gaussian trajectories are represented with learnable functions~\cite{Wu_2024_CVPR, 10657623, lin2024gaussian}. 
While more compact and efficient, these methods often struggle to capture complex or rapid motions, especially when relying solely on photometric cues. 

\subsection{Monocular Dynamic 3D Scene Reconstruction}

Monocular dynamic scene reconstruction is highly practical in real-world applications, but inherently ambiguous due to partial visibility of scene geometry and motion.
Several NeRF-based methods proposed to address this by incorporating object-specific priors such as human shape models~\cite{9879404, liu2021neural}, 
shadow modeling~\cite{d2nerf}, and facial expressions~\cite{9578714}. 
While effective, these methods generalize poorly beyond specific object categories. 
To improve generalization, geometric and physical priors such as depth, optical flow, and rigidity-based deformation constraints have been explored~\cite{du2021nerflow, Gao-ICCV-DynNeRF, tretschk2020nonrigid}, but monocular NeRFs still suffer from long training times and slow rendering.

Recent research has shifted toward explicit representations using 3D Gaussian Splatting, which offers faster and more efficient reconstruction. 
The progress in this area began with the work of Stearns \etal~\cite{stearns2024marbles}, who introduced isotropic Gaussian representations combined with motion trajectory blending from short sequences, guided by segmentation and depth priors. 
With advancements in point tracking, newer approaches propose to extract per-point trajectories from video and use them as motion bases shared across dynamic Gaussians~\cite{som2024}. 
The current state-of-the-art by Lei \etal~\cite{11094310} extends this idea with a Motion Scaffold Graph that connects trajectories into a spatio-temporal structure and refines them using rigidity and smoothness constraints.

Although these approaches achieve high-quality results, they remain limited by per-frame deformation modeling, leading to long training times and high memory usage. 
In contrast, we propose a compact and efficient representation for monocular dynamic scene reconstruction that achieves comparable quality to state-of-the-art methods 
while being significantly faster in both training and rendering.

\begin{figure*}[t]  %
\centering
\includegraphics[width=0.97\textwidth]{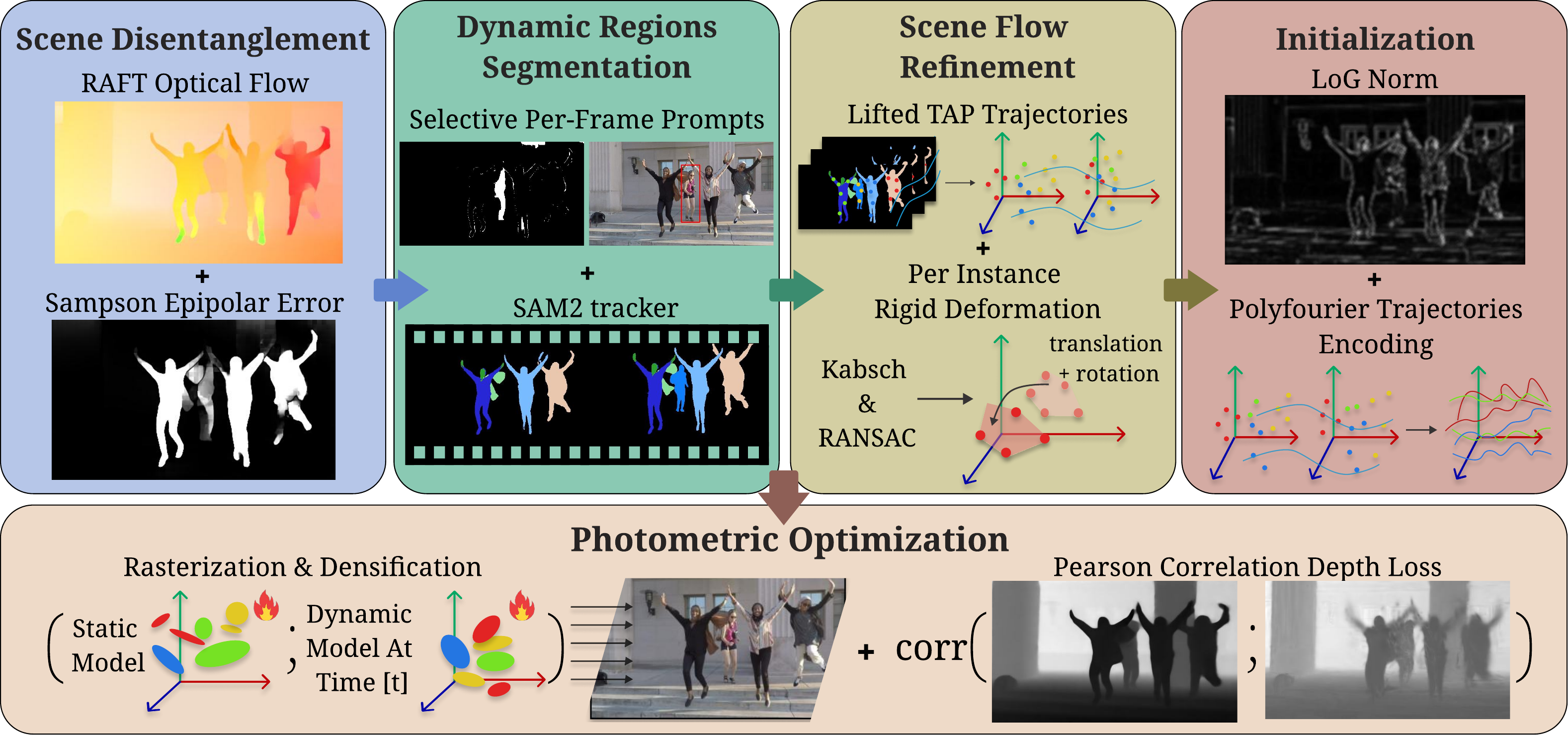}  %
\caption{\textbf{Overview of the MOSAIC-GS framework.} The pipeline begins with four pre-processing steps: 
(1) dynamic region detection using optical flow and epipolar error, (2) segmentation and tracking of dynamic instances, (3) scene flow estimation with per-instance 
rigidity refinement, and (4) initialization of static and dynamic Gaussians with Poly-Fourier encoded trajectories. The extracted parameters serve as initialization for the main photometric optimization phase, where separate Gaussian models are created for static and dynamic regions to improve parameter efficiency, while being jointly rasterized for rendering. 
A depth-based Pearson correlation loss further enhances geometric consistency and reconstruction fidelity.}
\label{fig:pipeline}
\end{figure*} 

\section{Method}

{\bf Overview.} Given a monocular video of a dynamic scene, along with camera intrinsics $\mathbf{K}$, extrinsics $[\mathbf{R} \mid \mathbf{t}]_t$, and per-frame depth maps $\mathbf{D_t}$ obtained from sensors or monocular depth estimation models~\cite{depthanything}, our pipeline accurately reconstructs dynamic 3D scenes while reducing training and rendering time.

Preliminary experiments show that photometric optimization is highly sensitive to scene initialization: without accurate motion estimates, most dynamic-region points are initially pruned. 
This leads to longer training times and reduced reconstruction quality, as the model must recover lost data over many iterations. Meuleman \etal~\cite{meuleman2025onthefly} similarly highlight that proper initialization accelerates convergence and improves visual quality. 
To prevent this data loss, we argue that the initialization should include richer information, rather than relying solely on Gaussian position and color as in most prior works~\cite{kerbl3Dgaussians, 10656774, lin2024gaussian}.

To achieve this, we propose an advanced pre-processing pipeline with four
main stages: (1) dynamic region detection, (2) segmentation and tracking of dynamic instances, (3) scene flow estimation and per-instance refinement using rigidity constraints, and finally (4) initialization of static and dynamic regions. In the last stage, the extracted scene flow is encoded as time-dependent Poly-Fourier curves $\mathbf{F}_t^i$ for compact trajectory representation, and initialization points are sampled adaptively to match the density of detail in each region.

The initialized Gaussians are then optimized and densified during the photometric optimization phase, regularized by a depth-based supervision loss. 
A full overview of the framework is shown in Fig.~\ref{fig:pipeline}, and each stage is described in detail below.

\subsection{Dynamic Regions Detection}
To identify dynamic objects in the scene, we leverage dense optical flow in combination with epipolar geometry constraints. Instead of relying on full 3D reconstruction errors, which can be unreliable due to noisy depth maps, we employ the Sampson epipolar error. 

Given two consecutive frames $\mathbf{I_t}$ and $\mathbf{I_{t+1}}$, 
we first compute the dense optical flow $\mathbf{u}_t: \Omega \to \mathbb{R}^2$ using RAFT~\cite{raft}. 
Following Liu \etal~\cite{liu2023robust}, we compute the Sampson epipolar error for each pixel correspondence $\mathbf{x} \in \Omega$ using known camera extrinsics $[\mathbf{R} \mid \mathbf{t}]_t$, $[\mathbf{R} \mid \mathbf{t}]_{t+1}$ and intrinsics $\mathbf{K}$,:

\begin{equation}
    e_{\text{epi}}(\mathbf{x}) = \frac{(\mathbf{x}'^\top \mathbf{F} \mathbf{x})^2}{(\mathbf{F}\mathbf{x})_1^2 + (\mathbf{F}\mathbf{x})_2^2 + (\mathbf{F}^\top \mathbf{x}')_1^2 + (\mathbf{F}^\top \mathbf{x}')_2^2},
\end{equation}
where $\mathbf{F}$ is the fundamental matrix derived from $[\mathbf{R} \mid \mathbf{t}]_t$, $[\mathbf{R} \mid \mathbf{t}]_{t+1}$ and $\mathbf{K}$, and $\mathbf{x}'$ is the corresponding pixel in the next frame. Pixels with high epipolar error correspond to regions that cannot be explained by camera motion alone,
which may arise from true scene motion, inaccuracies in optical flow estimation, or errors in camera parameters. To filter out false positives, we consider dynamic only those regions where
the epipolar error satisfies $e_{\text{epi}}(\mathbf{x}) > \tau_{\text{epi}}$.

\subsection{Segmentation and Tracking of Dynamic Instances}

Relying solely on epipolar error thresholding can be insufficient in certain scenarios. To improve the detection, tracking, and segmentation of dynamic regions across all frames, we incorporate the prompt-based semantic segmentation and tracking model SAM2~\cite{ravi2024sam2}.

For each frame, we first extract bounding boxes around dynamic regions detected in the previous step.
If no dynamic objects have been detected in earlier frames, these boxes are used directly as prompts for the segmentation model~\cite{ravi2024sam2}. Otherwise, masks of previously detected and tracked instances are first subtracted from the dynamic regions to prevent duplicate instance tracking. Bounding boxes around the remaining regions are then used as prompts to generate accurate instance masks $\mathbf{M_t^j}$ for each new dynamic object $j$.
Masks with low confidence scores $p_{\text{conf}}(\mathbf{M_t^j}) < \tau_{\text{mask}}$, are filtered out to remove unreliable detections.

\begin{figure}[t]
\begin{center}
\includegraphics[width=0.97\linewidth]{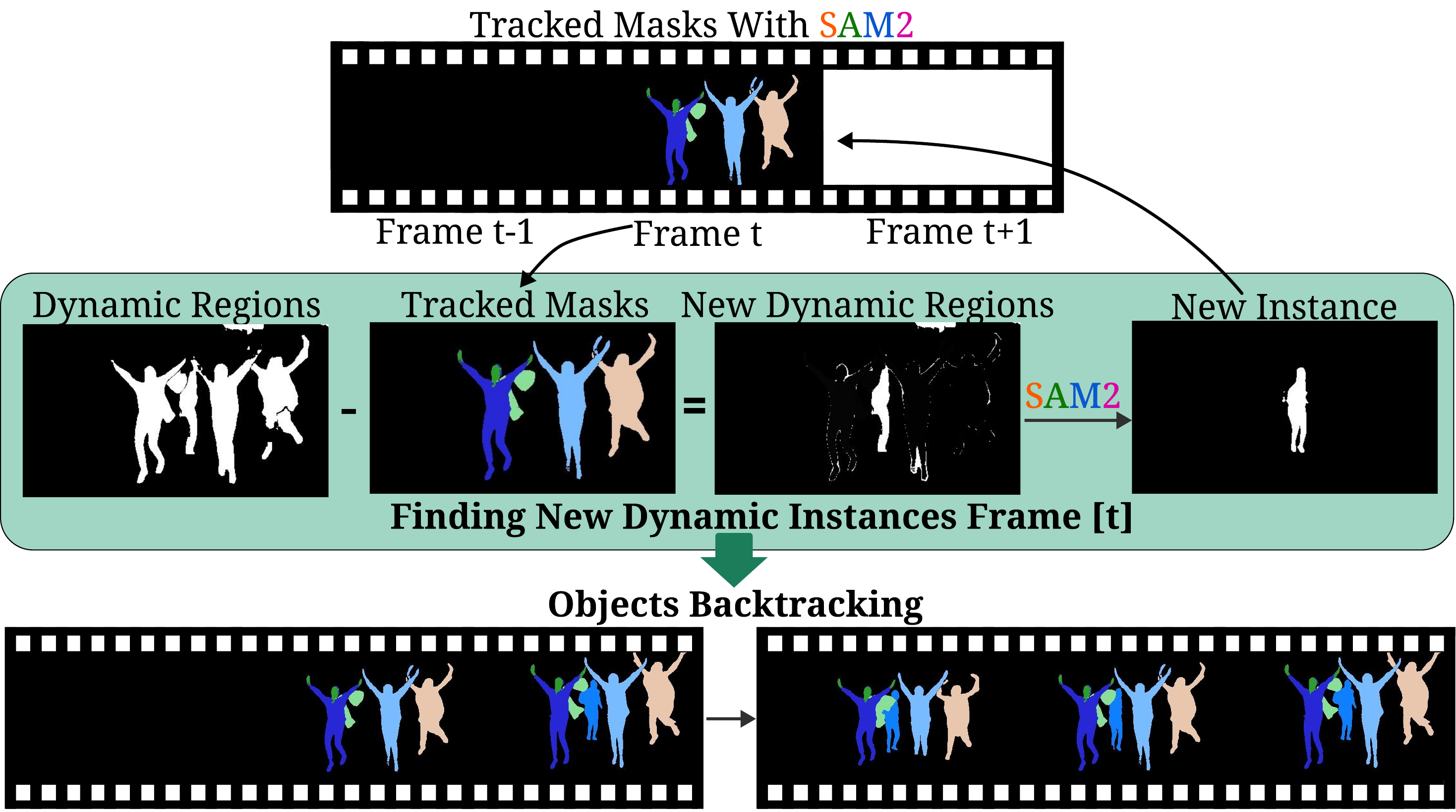}
\end{center}
   \caption{\textbf{Dynamic object detection and tracking.} Masks of previously detected objects are subtracted from dynamic regions to identify new instances. The remaining regions are used as prompts for SAM2~\cite{ravi2024sam2} to generate segmentation masks, which are tracked forward through all frames. A reverse propagation pass extends object masks to earlier frames where the object was initially static.}
\label{fig:sam2_tracking}
\end{figure}

High-confidence masks are added to the tracker and propagated forward at fixed intervals, reducing the computational overhead of re-tracking frames when new dynamic objects appear. After processing all frames, we obtain a set of tracked dynamic object masks $\{\mathbf{M_t^j}\}_{t=1}^T$ for the entire video. To account for objects that were static in earlier frames, we perform a reverse propagation pass, extending object masks to previous frames. 
A visualization of this process is shown in Fig.~\ref{fig:sam2_tracking}.

\subsection{Scene Flow Estimation and Refinement}
To extract scene flow for dynamic regions, we leverage recent advances in point tracking~\cite{doersch2024bootstap,cotracker}, 
following prior works~\cite{11094310,som2024}. 
We randomly sample $N_p = 10,000$ query points $\mathbf{q_t^i}$ within dynamic regions across all $T$ frames and track them through the video, where $i$ denotes the point index 
and $t$ the index of a frame where the point has been queried.

Each trajectory $\{\mathbf{p_t^i}\}_{t=1}^T$ is assigned to a dynamic object $j$ based on majority overlap with segmentation masks $\mathbf{M_t^j}$, producing per-object 3D point sets $\{\mathcal{P}_j\}_{j=1}^J$.
Trajectories primarily associated with the static background are discarded. Combining segmentation and tracking resolves ambiguities from occlusions and imperfect masks, producing more temporally consistent instance identifiers. 

The 2D trajectories are then lifted to 3D using depth maps $\mathbf{D_t}$, camera intrinsics $\mathbf{K}$, and extrinsics $[\mathbf{R} \mid \mathbf{t}]_t$ to produce scene flow $\{ \mathbf{P_t^i} \mid i = 1, \dots, N_p; \ t = 1, \dots, T \}$.
To reduce noise and fill unobserved regions, we refine the flow of each dynamic object using rigid transformations. 
For object $j$, we consider points 
$\{ (\mathbf{P_t^i}, \mathbf{P_{t+1}^i}) \mid i \in \mathcal{P}_j, \; v_t^i = 1, \; v_{t+1}^i = 1 \}$, where $v_t^i$ indicates visibility at frame $t$, and estimate the
rigid transformation $(\mathbf{R_t^j}, \mathbf{t_t^j})$ that best aligns them using the Kabsch algorithm~\cite{kabsch1976solution} with RANSAC~\cite{fischler1981random}.
Refined positions at $t+1$ are then computed as:
\begin{equation}
\mathbf{P}_{t+1}^i := \mathbf{R}_t^j \mathbf{P}_t^i + \mathbf{t}_t^j, \quad
v_{t+1}^i := 1 \;\big|\; v_t^i = 1
\end{equation}
After processing all frames forward, we apply the same refinement backward to update points that were not observed in earlier frames but whose corresponding objects were visible at those frames.
Remaining points with $v_t^i = 0$ are  interpolated using the nearest frames $t' < t$ and $t'' > t$, where $v_{t'}^i = 1$ and $v_{t''}^i = 1$.

\subsection{Initialization of Static and Dynamic Regions}
To provide a compact representation of the scene we propose representing scene dynamics as time-dependent Poly-Fourier curves, following Lin \etal\cite{lin2024gaussian}.
However, instead of learning these curves' coefficients during the photometric optimization phase, we propose initializing them directly from the refined scene flow
obtained in the previous step. To achieve this, we are solving this system of linear equations for each trajectory $\{\mathbf{P_t^i}\}_{t=1}^T$:

\begin{equation}
A x = y
\label{eq:Ax=y}
\end{equation}

\[
\begin{aligned}
\text{where:}\\
A &= 
\begin{bmatrix}
\phi(t_1) & \phi(t_2) & \cdots & \phi(t_T)
\end{bmatrix}^\top, \\[1mm]
\phi(t) &= [\,1,\, t,\, t^2,\, \sin(\omega t),\, \cos(\omega t)\,]^\top, \\[1mm]
y &= [\,P_i(t_1),\, P_i(t_2),\, \ldots,\, P_i(t_T)\,]^\top, \\[1mm]
x &= [\,a_0,\, a_1,\, a_2,\, \ldots,\, b_1,\, c_1,\, \ldots\,]^\top.
\end{aligned}
\]

The resulting coefficients $\mathbf{x}$ then compactly encode each trajectory over time and are used to initialize the deformation parameters of dynamic Gaussians.
In addition to motion, we initialize each Gaussian’s color and scale to match the local appearance and level of details. Specifically, for each trajectory query point $\mathbf{q}_t^i$, we extract its color from the corresponding pixel in the RGB frame $\mathbf{I}_t$.
To estimate the Gaussian scale, we evaluate the Laplacian of Gaussian (LoG) norm on $\mathbf{I}_t$ and incorporate its depth map $\mathbf{D}_t$ as proposed by Meuleman \etal~\cite{meuleman2025onthefly}. The extracted Poly-Fourier coefficients, colors, and scales together initialize the mean position, deformation parameters, color, and scale of each dynamic Gaussian.

For static regions, we adopt the sampling strategy of Meuleman \etal~\cite{meuleman2025onthefly}. To ensure broad scene coverage while avoiding oversampling of flat areas, sampling is performed only on a subset of temporally spaced training frames.

\begin{table*}[t]
\begin{center}
\resizebox{\textwidth}{!}{ 
\begin{tabular}{l|ccccccc|c}
\toprule
\textbf{PSNR$\uparrow$ / LPIPS$\downarrow$} & Apple & Block & Spin & Paper Windmill & Space-Out & Teddy & Wheel & Mean \\
\midrule
Dyn. Gaussians\cite{10550869} & 7.65 / 0.766 & 7.55 / 0.684 & 8.08 / 0.651 & 6.24 / 0.729 & 6.79 / 0.733 & 7.41 / 0.690 & 7.28 / 0.593 & 7.29 / 0.692 \\
4D Gaussians\cite{10656774}   & 15.41 / 0.450 & 11.28 / 0.633 & 14.42 / 0.339 & 15.60 / 0.297 & 14.60 / 0.372 & 12.36 / 0.466 & 11.79 / 0.436 & 13.64 / 0.428 \\
Marbles\cite{stearns2024marbles} & 17.70 / 0.492 & 17.42 / 0.384 & 18.88 / 0.428 & 17.04 / 0.394 & 15.94 / 0.435 & 13.95 / 0.442 & 16.14 / 0.351 & 16.72 / 0.418 \\
Gaussian Flow\cite{10658506} & 18.37 / \textbf{0.322} & 15.62 / 0.352 & 16.66 / 0.302 & 18.14 / 0.217 & 17.12 / 0.286 & 13.48 / 0.376 & 14.18 / 0.323 & 16.22 / 0.311 \\
Shape of Motion\cite{som2024} & 18.57 / 0.341 & 17.41 / \underline{0.323} & 19.35 / 0.247 & 18.14 / 0.225 & 16.85 / 0.324 & 13.69 / 0.38 & 17.21 / 0.23 & 17.32 / 0.295 \\
MoSca\cite{11094310} & \textbf{19.4} / 0.34 & \textbf{18.06} / 0.33 & \textbf{21.31} / \textbf{0.19} & \textbf{22.34} / \underline{0.15} & \textbf{20.48} / \underline{0.26} & \textbf{15.47} / \underline{0.35} & \textbf{18.17} / \underline{0.23} & \textbf{19.32} / \underline{0.264} \\
Ours & \underline{18.92} / \underline{0.328} & \underline{17.5} / \textbf{0.29} & \underline{20.28} / \underline{0.233} & \underline{20.31} / \underline{0.15} & \underline{19.96} / \textbf{0.239} & \underline{14.42} / \textbf{0.343} & \underline{17.40} / \textbf{0.204} & \underline{18.4} / \textbf{0.255} \\
\bottomrule
\end{tabular}}
\end{center}
\caption{Per-scene quantitative results on the DyCheck~\cite{gao2022dynamic} dataset. Best results are in \textbf{bold}, second best are \underline{underlined}. 
MOSAIC-GS achieves comparable performance to the state-of-the-art MoSca~\cite{11094310} method, while being significantly faster in both training and rendering and achieves
a new state-of-the-art in terms of LPIPS for this dataset.}
\label{tab:dycheck_comparison}
\end{table*}

\subsection{Photometric Optimization}

The proposed MOSAIC-GS framework represents the scene using two distinct sets of 3D Gaussians: a static component $\mathcal{G}_s$ and a dynamic component $\mathcal{G}_d$.  
Each component is initialized separately and follows a distinct densification strategy to account for their differing parameter complexity.
Formally, the full scene representation is defined as $\mathcal{G} = \mathcal{G}_s \cup \mathcal{G}_d$.

Each Gaussian in the static set $\mathcal{G}_s$ is parameterized by its mean position 
$\boldsymbol{\mu} \in \mathbb{R}^3$, rotation quaternion $\mathbf{q} \in \mathbb{H}$, 
scale $\mathbf{s} \in \mathbb{R}^3$, opacity $\alpha \in [0,1]$, and color coefficients $\mathbf{c} \in \mathbb{R}^{3 \times (B+1)^2}$, where $B$ denotes the degree of spherical harmonics.

Dynamic Gaussians in $\mathcal{G}_d$ share the same canonical parameters but additionally store 
time-dependent deformation parameters encoded as Poly-Fourier coefficients 
$\{\mathbf{a}_k\}_{k=1}^{d_{\text{pol}}}$, $\{\mathbf{b}_k, \mathbf{c}_k\}_{k=1}^{d_{\text{Fourier}}}$ representing offsets to the Gaussian mean and rotation.
Position and rotation deformations are modeled separately, with positional parameters initialized 
from the Poly-Fourier coefficients estimated during the pre-processing stage. 
The position offset at time $t$ is given by:
\begin{equation}
\Delta \boldsymbol{\mu}(t) = \sum_{k=1}^{d_p} a_k t^k + \sum_{k=1}^{d_F} \Big( b_k \cos(k \omega t) + c_k \sin(k \omega t) \Big),
\end{equation}
and the mean position becomes $\boldsymbol{\mu}(t) = \boldsymbol{\mu}_0 + \Delta \boldsymbol{\mu}(t)$.

For rotation, instead of directly adding time-dependent offsets to the base quaternion 
as in Gaussian Flow~\cite{lin2024gaussian}, we convert the Poly-Fourier output into a valid rotation 
by adding the identity quaternion and normalizing it to form a unit quaternion $\Delta \mathbf{q}(t)$. 
The final rotation is then obtained as:
\begin{equation}
\mathbf{q}(t) = \Delta \mathbf{q}(t) \otimes \mathbf{q}_0,
\end{equation}
where $\mathbf{q}_0$ denotes the canonical rotation and $\otimes$ indicates quaternion multiplication.
This design enforces consistent rotation parameterization and avoids artifacts from invalid quaternion updates.

Unlike Gaussian Flow~\cite{lin2024gaussian}, we do not model time-varying color deformations, 
which improves compactness and prevents the model from compensating motion inaccuracies 
through artificial color changes.

The overall loss function is defined as:
\begin{equation}
\mathcal{L} = (1 - \lambda_{\text{ssim}})\, \mathcal{L}_{\text{L1}} 
+ \lambda_{\text{ssim}}\, \mathcal{L}_{\text{SSIM}} 
+ \lambda_{\text{depth}}\, \mathcal{L}_{\text{depth}},
\end{equation}
where $\mathcal{L}_{\text{L1}}$ and $\mathcal{L}_{\text{SSIM}}$ denote photometric losses, and $\mathcal{L}_{\text{depth}}$ enforces geometric consistency with reference depth maps. To handle temporally inconsistent depth scales, $\mathcal{L}_{\text{depth}}$ is implemented as a Pearson correlation loss~\cite{FSGS, liu2025modgs}, which preserves relative geometry while remaining invariant to absolute scale variations. Further implementation details are provided in supplementary materials. 

\section{Experiments}
\textbf{Datasets.}  
We evaluate MOSAIC-GS on standard monocular dynamic scene reconstruction benchmarks, 
including the \textit{iPhone DyCheck}~\cite{gao2022dynamic} and \textit{NVIDIA Dynamic Scene} dataset~\cite{9156445}. The \textit{iPhone DyCheck} dataset contains seven casually captured dynamic scenes recorded with a moving monocular camera, 
each with up to 500 frames~\cite{gao2022dynamic}. Each scene provides RGB images, camera intrinsics/extrinsics, noisy LiDAR depth maps, and ground-truth covisibility masks, along with one or two static cameras for novel-view evaluation.

Two distinct versions of the \textit{NVIDIA Dynamic Scene} dataset exist:  
(1) the \textit{original} release by Yoon et al.~\cite{9156445}, and 
(2) the modified \textit{Gaussian Marbles} variant introduced by Stearns et al.~\cite{stearns2024marbles}. The \textit{original} dataset~\cite{9156445} includes seven dynamic sequences captured by twelve static cameras. Monocular input is simulated by selecting one frame from each consecutive camera, yielding twelve spatially and temporally sparse frames. Evaluation uses images from the first camera as reference views. The \textit{Gaussian Marbles} version~\cite{stearns2024marbles} contains seven scenes with 100–200 frames each, captured by four static cameras. One camera provides monocular input, while the others serve as evaluation views.

This distinction has been overlooked in prior works\cite{11094310}, leading to inconsistent comparisons. To ensure fairness, we report results on both versions and compare with methods evaluated under the corresponding settings.

\subsection{Novel-View Synthesis}

\textbf{DyCheck~\cite{gao2022dynamic}.}  
Following MoSca~\cite{11094310}, we employ the BootsTAPIR tracker~\cite{doersch2024bootstap} to extract dense point trajectories across frames.
MOSAIC-GS focuses exclusively on novel-view synthesis and does not include a camera pose refinement stage. 
In contrast, prior works often perform additional pose optimization to compensate for noisy ground-truth poses~\cite{som2024, 11094310}. 
For fair comparison, we adopt the refined poses extracted by MoSca~\cite{11094310}.

\begin{figure}[t]
\begin{center}
\includegraphics[width=\linewidth]{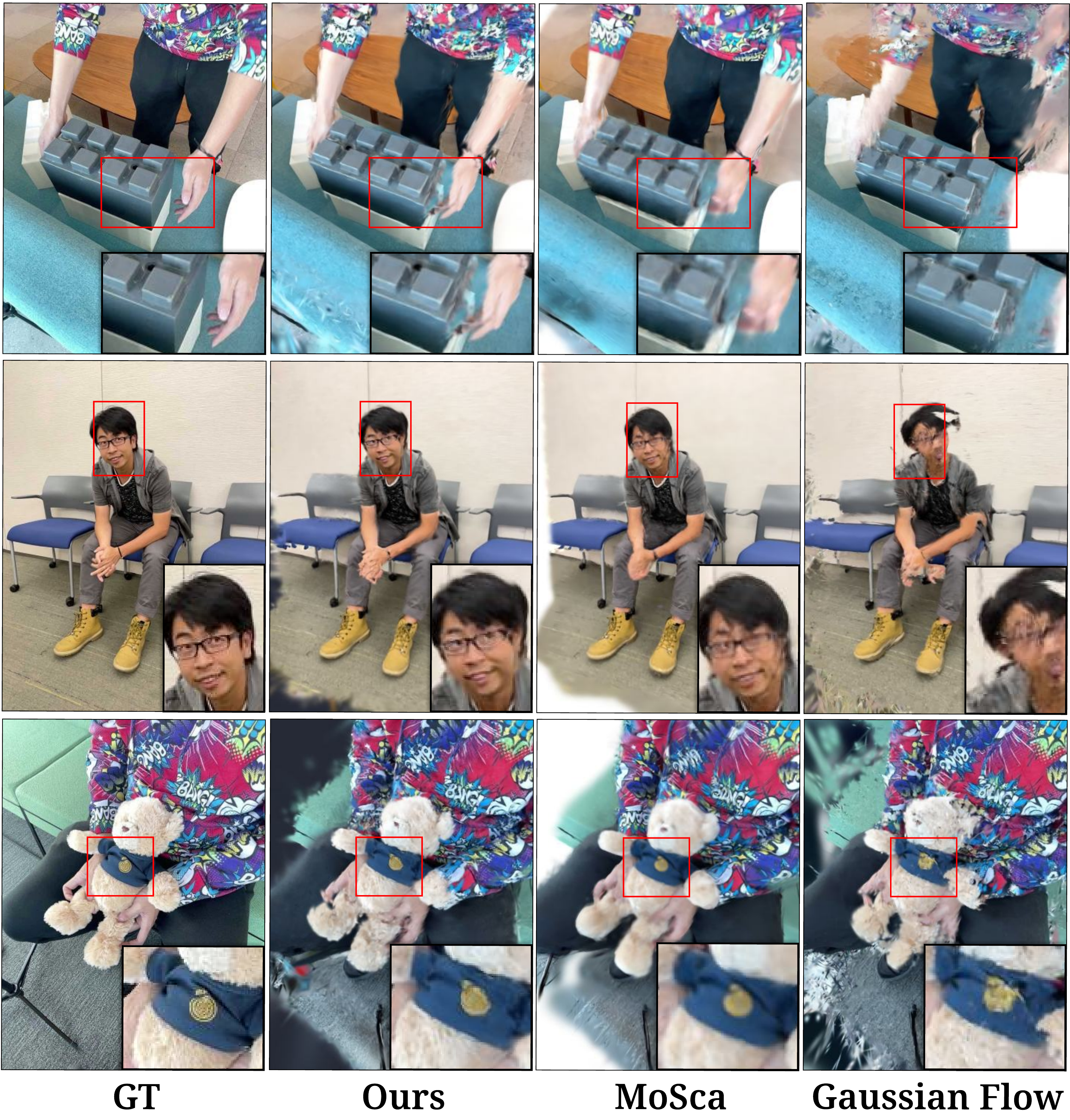}
\end{center}
   \caption{Qualitative comparison of novel view synthesis on the DyCheck~\cite{gao2022dynamic} dataset. Our method produces sharper details in dynamic regions compared to prior approaches
   resulting in higher perceptual similarity with the ground truth evaluation view.}
\label{fig:dycheck_comparison}
\end{figure}

Table~\ref{tab:dycheck_comparison} reports quantitative results comparing MOSAIC-GS with state-of-the-art monocular dynamic scene 
reconstruction methods, evaluated using PSNR and LPIPS. MOSAIC-GS achieves competitive performance, outperforming all prior methods 
except MoSca~\cite{11094310} in PSNR, while setting a new state-of-the-art in LPIPS across most scenes and on average. Notably, these results 
are obtained with substantially lower training and rendering times as shown in Table~\ref{tab:timings}, enabled by our compact scene representation, static–dynamic scene disentanglement, and accurate motion initialization, 
which reduce the number of optimization steps required for reconstruction. Qualitative comparisons are shown in Fig.~\ref{fig:dycheck_comparison}. Results for Gaussian Marbles are omitted due to the unavailability of their preprocessed dataset.
MOSAIC-GS reconstructs finer details and produces sharper renderings in dynamic regions, resulting in noticeably better perceptual quality.
While this level of detail may slightly reduce PSNR, the improvement is better reflected by perceptual metrics such as LPIPS, which align more closely with visual quality.

\begin{figure}[t]
\begin{center}
\includegraphics[width=0.99\linewidth]{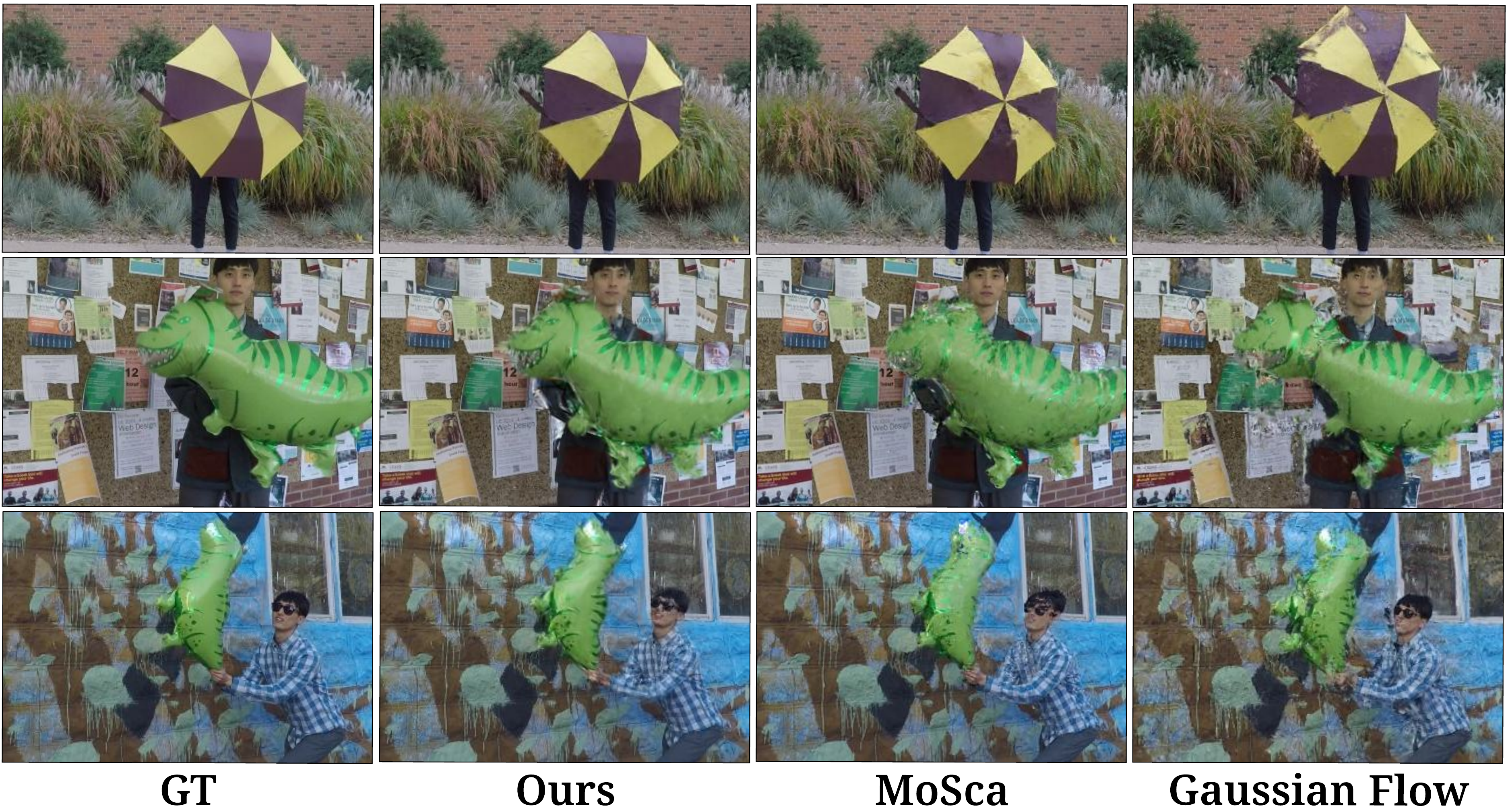}
\end{center}
   \caption{Qualitative comparison of novel view synthesis on the NVIDIA Dynamic Scene (original) dataset. The proposed method better captures motion and 
   preserves objects' geometry when observed from different viewpoints.}
\label{fig:nvidia_comparison}
\end{figure}

\textbf{NVIDIA Dynamic Scene Dataset~\cite{9156445}.}  
Table~\ref{tab:nvidia_comparison} presents PSNR and LPIPS scores averaged across all scenes. Similarly to DyCheck, MOSAIC-GS achieves 
second-best PSNR after MoSca~\cite{11094310} and attains the best LPIPS among all compared methods.
As illustrated in Fig.~\ref{fig:nvidia_comparison}, MOSAIC-GS more faithfully captures motion and preserves object geometry when observed from novel viewpoints, 
resulting in visually more coherent reconstructions.

\textbf{NVIDIA Dynamic Scene Dataset (modified~\cite{stearns2024marbles}).}  
Since this dataset does not provide reliable covisibility masks, prior works evaluated reconstruction quality on the entire image, often applying post-processing to fill unobserved regions~\cite{stearns2024marbles}. 
We argue that this approach is less accurate than restricting evaluation to covisible regions. 
Accordingly, we generate covisibility masks analytically from depth maps and camera parameters and limit quantitative evaluation to these areas.

Table~\ref{tab:nvidia_comparison_marbles} presents results using this covisibility-based protocol, re-evaluating all prior approaches 
using the same procedure. On this dataset MOSAIC-GS achieves state-of-the-art performance in both PSNR and LPIPS. Additional qualitative 
comparisons are provided in the supplementary materials.

\begin{table}[t]
\begin{center}
\resizebox{\columnwidth}{!}{
\begin{tabular}{l|c|c}
\toprule
\textbf{Methods} & \textbf{Training Time}$\downarrow$ & \textbf{Rendering Speed}$\uparrow$ \\
\midrule
Gaussian Marbles* & 5-9h & \textbf{200 FPS} \\
Gaussian Flow & \underline{23 min} & 52 FPS \\
MoSca & 50 min & 38 FPS \\
Ours & \textbf{10.5 min} & \underline{180 FPS} \\
\bottomrule
\end{tabular}}
\end{center}
\caption{Comparison of training time and rendering speed averaged across the DyCheck~\cite{gao2022dynamic} dataset. *: The values for Gaussian Marbles are taken from the original paper\cite{stearns2024marbles}. 
All other methods were benchmarked on one NVIDIA RTX 4090 GPU. The training time reports the total time including pre-processing, the photometric optimization takes approximately 5 minutes.}
\label{tab:timings}
\end{table}

\begin{table}[t]
\begin{center}
\resizebox{0.6\columnwidth}{!}{
\begin{tabular}{l|cc}
\toprule
\textbf{Method} & \textbf{PSNR}$\uparrow$ & \textbf{LPIPS}$\downarrow$ \\
\midrule
Casual-FVS~\cite{lee2023casual-fvs} & 24.57 & 0.081 \\
MonoNeRF~\cite{mononerf} & 25.62 & 0.106 \\
CTNeRF~\cite{ctnerf} & 26.13 & 0.082 \\
RoDynRF~\cite{liu2023robust} & 25.89 & \underline{0.067} \\
MoSca~\cite{11094310} & \textbf{26.72} & 0.070 \\
\textbf{Ours} & \underline{26.26} & \textbf{0.060} \\
\bottomrule
\end{tabular}}
\end{center}
\caption{Comparison on NVIDIA Dataset (original)\cite{9156445}, averaged over all scenes.}
\label{tab:nvidia_comparison}
\end{table}

\begin{table}[t]
\begin{center}
\resizebox{\columnwidth}{!}{
\begin{tabular}{l|cc|c}
\toprule
\textbf{Method} & \textbf{PSNR}$\uparrow$ & \textbf{LPIPS}$\downarrow$ & \textbf{Training Time} \\
\midrule
Gaussian Flow~\cite{10658506} & 21.89 & 0.095 & 18 min \\
Gaussian Marbles~\cite{stearns2024marbles} & 23.68 & \underline{0.069} & 3.5-8 h \\
Ours & \textbf{23.79} & \underline{0.069} & \textbf{7 min} \\
\midrule
\end{tabular}}
\end{center}
\caption{Results averaged over all scenes on the Gaussian Marbles variant of the NVIDIA dataset~\cite{stearns2024marbles}. Our method achieves state-of-the-art quality with a fraction of the training time of Dynamic Gaussian Marbles~\cite{stearns2024marbles}.}
\label{tab:nvidia_comparison_marbles}
\end{table}

\subsection{Ablation Study}
We conduct an ablation study on the DyCheck dataset~\cite{gao2022dynamic} to evaluate the impact of key components of the MOSAIC-GS framework. 
Specifically, we investigate the effects of: 
(1) initialization of mean deformation coefficients for dynamic Gaussians,
(2) scene flow refinement using rigid transformations,
(3) scene disentanglement into static and dynamic Gaussians,
(4) different degrees of Fourier encoding for motion representation, compared to the default degree of 32, and
(5) depth supervision during photometric optimization.
 
As shown in Table~\ref{tab:ablation_study}, each component of our approach improves the reconstruction quality. 
The largest drop occurs when the scene disentanglement module is removed, leading to poor initialization of static regions and incorrect scene flow refinement. This also significantly increases computational cost, as every Gaussian must store deformation parameters. Removing the initialization of deformation parameters for dynamic Gaussians also substantially reduces visual quality, highlighting the importance of accurate motion priors. 
Additional scene flow refinement further improves results, especially in scenes with complex motion, 
where point tracking benefits from extra refinement.  

\begin{figure}[t]
\begin{center}
\includegraphics[width=0.99\linewidth]{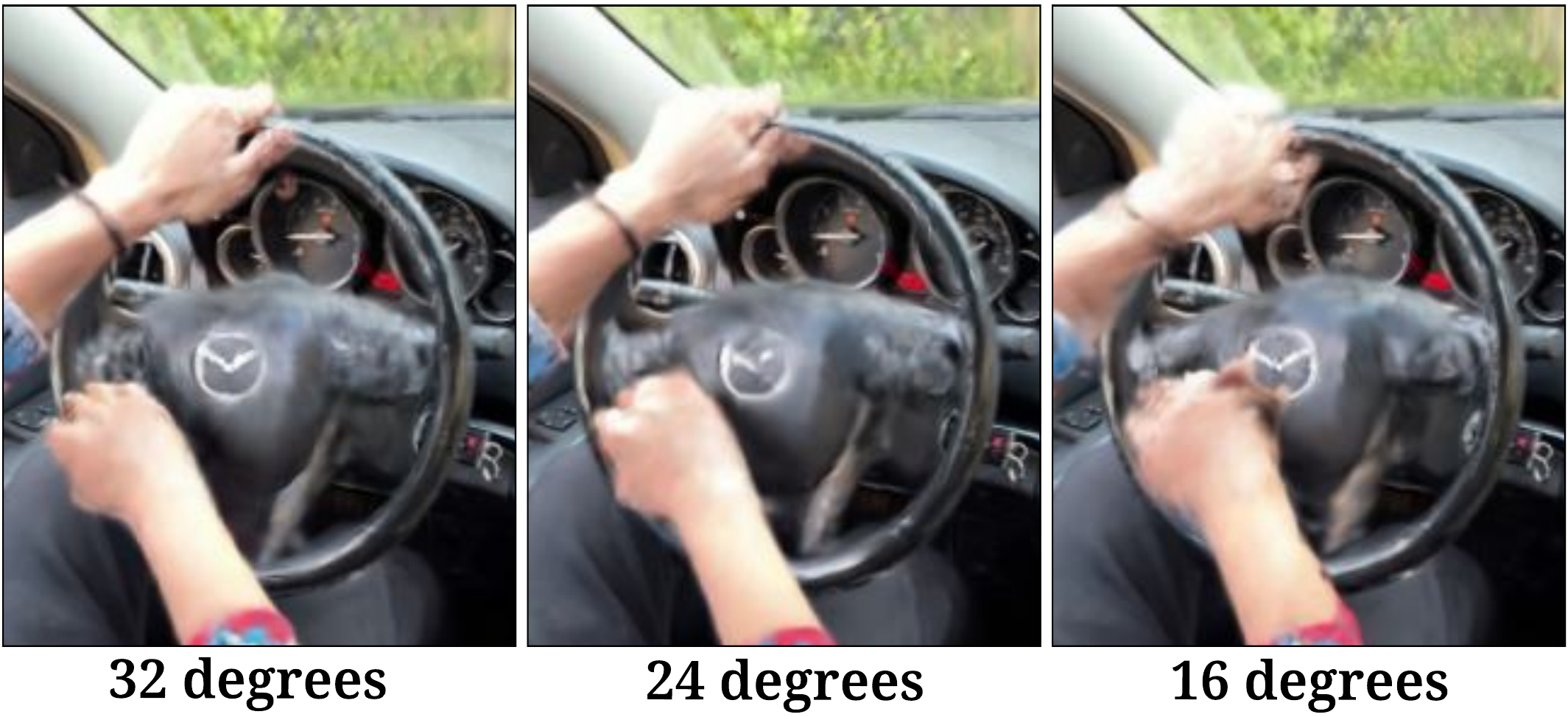}
\end{center}
   \caption{Effect of Fourier degree on the reconstruction of fine details in dynamic regions.}
\label{fig:fourier_degree_comparison}
\end{figure}

The choice of Fourier degree has a relatively small overall effect, though higher degrees better capture complex motion in detailed, fast-moving scenes like \textit{Wheel} (Fig.~\ref{fig:fourier_degree_comparison}). Fourier degrees higher than 32 may offer marginal gains in certain cases, but they sacrifice compactness and were therefore not explored further. Lower degrees, on the other hand, provide a more compact motion representation, which can be beneficial in resource-constrained settings. 

Finally, adding the depth supervision term further improves reconstruction quality by enhancing both photometric accuracy and geometric consistency. 

\begin{table}[t]
\begin{center}
\resizebox{\columnwidth}{!}{
\begin{tabular}{l|ccc|c}
\toprule
\textbf{Components} & \textbf{mPSNR}$\uparrow$ & \textbf{mSSIM}$\uparrow$ & \textbf{mLPIPS}$\downarrow$ & \textbf{Time}$\downarrow$ \\
\midrule
\textbf{Full model} & \textbf{18.40} & \textbf{0.668} & \textbf{0.255} & 5.06 \\
No deformations initialization & 16.99 & 0.641 & 0.298 & 5.41 \\
No flow refinement & 18.11 & 0.663 & 0.265 & 4.83 \\
No scene disentanglement & 14.65 & 0.521 & 0.456 & 8.26 \\
Fourier degree 24 & 18.37 & 0.667 & 0.261 & 4.91 \\
Fourier degree 16 & 18.29 & 0.665 & 0.265 & \textbf{4.68} \\
No depth supervision & 18.13 & 0.665 & 0.264 & 4.87 \\
\bottomrule
\end{tabular}}
\end{center}
\caption{Ablation study on different components of the system on the DyCheck dataset~\cite{gao2022dynamic}. The \textit{Time} column reports the training duration of the photometric optimization stage (in minutes). }
\label{tab:ablation_study}
\end{table}

\subsection{Additional Applications}
MOSAIC-GS additionally allows temporally consistent scene segmentation 
and dynamic scene editing without any extra computational cost. During pre-processing, temporal refinement of mask identifiers with point tracking ensures consistent segmentation of dynamic objects over time, avoiding identity swaps that occur with simple 3D reprojection of masks. 
By assigning segmentation IDs to dynamic Gaussians during initialization, the densification and optimization steps naturally produce temporally consistent segmentation of dynamic regions as a by-product of the reconstruction process.

\begin{figure}[t]
\begin{center}
\includegraphics[width=\linewidth]{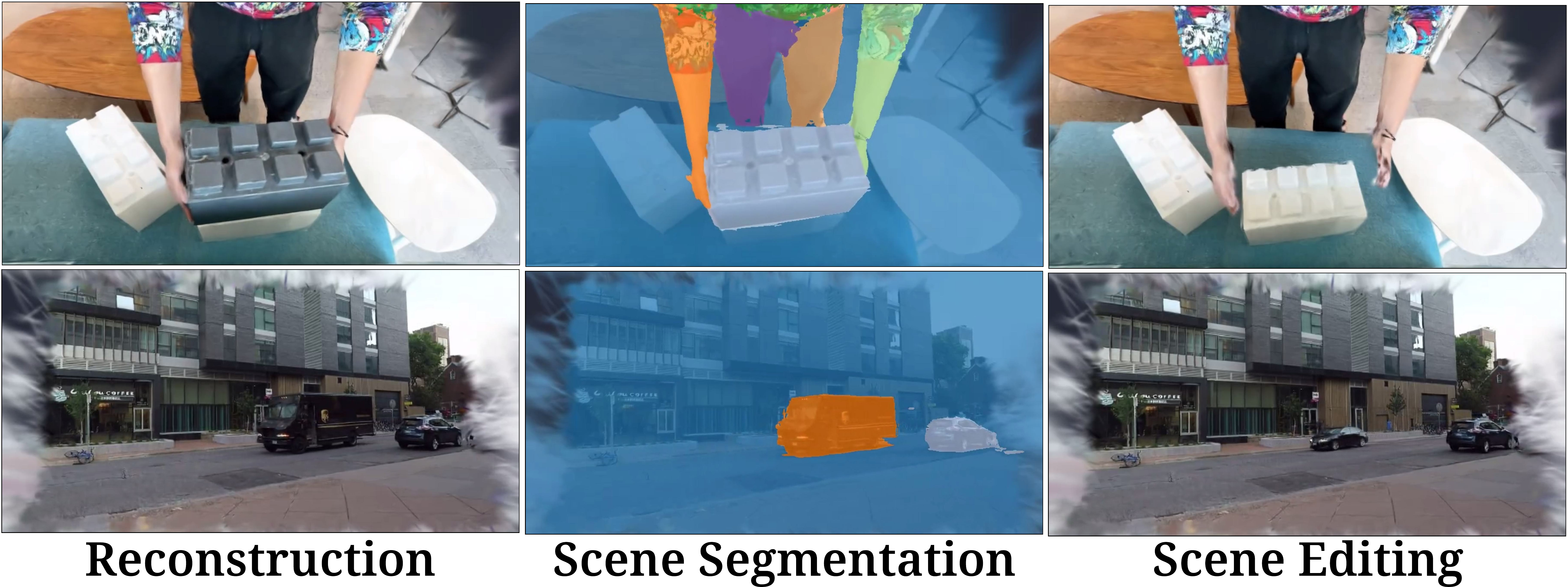}
\end{center}
   \caption{Examples of scene segmentation and editing enabled by the instance IDs stored in dynamic Gaussians. In the top image, the gray cube has been removed from the scene, while in the bottom image, the truck has been removed.}
\label{fig:add_applications}
\end{figure}

This allows us not only to remove all dynamic objects from the scene, or, conversely, render only the dynamic objects on a blank background, but also to perform several editing operations,  such as changing object color, adjusting motion or position, 
or simply isolating them from the rest of the scene during rendering. Examples of these editing results are shown in Fig.~\ref{fig:add_applications}

\section{Limitations}
While MOSAIC-GS achieves efficient and accurate monocular dynamic scene reconstruction, it has several limitations.  
First, the approach relies heavily on the quality of initial segmentation masks and scene flow, which may inherit errors from the external models.  
Second, the scene flow refinement may be insufficient, particularly when an entire dynamic object is invisible from the training viewpoint.  
Further discussion of limitations and future research directions is provided in the supplementary materials.

\section{Conclusion}
In this work, we presented MOSAIC-GS, a framework for efficient and accurate monocular dynamic scene reconstruction using 3D Gaussian Splatting. 
Experiments on standard benchmarks show that our method achieves the state-of-the-art training and rendering efficiency, along with superior perceptual quality (LPIPS) in novel-view synthesis. These improvements stem from an efficient pre-processing stage, including scene disentanglement, scene flow refinement, and compact motion encoding using Poly-Fourier curves. We believe this work takes a meaningful step toward fast dynamic scene reconstruction with high visual fidelity, opening up new possibilities for applications in AR/VR, robotics, and 3D content creation.

\bigskip

\textbf{Acknowledgment.} This work was partially supported as a part of NCCR Automation, a National Centre of Competence in Research, funded by the Swiss National Science Foundation (grant number 51NF40\textunderscore225155). This work is also supported by Swiss Federal Railways (SBB) through the ETH Mobility Initiative and ETHAR, the ETH Augmented Reality Research Hub.

{\small
\bibliographystyle{ieee_fullname}
\bibliography{egbib}
}

\end{document}

\title{Supplementary Material for \enquote{MOSAIC-GS: Monocular Scene Reconstruction via Advanced Initialization for Complex Dynamic Environments}}

\maketitle

\section{Implementation details}

\subsection{Dynamic Objects Tracking and Scene Flow Extraction}

To reduce initialization time, we extract dynamic object masks and their temporal tracks using SAM2~\cite{ravi2024sam2} only on a subset of frames: up to 150 frames for iPhone DyCheck~\cite{gao2022dynamic} sequences and 30 frames for the NVIDIA Dynamic Scene datasets~\cite{9156445, stearns2024marbles}. Sufficient temporal spacing between sampled frames is important, as too-small gaps may cause true object motion to be mistaken for noise during epipolar error thresholding. Subsampling naturally enforces this separation.
We use a Sampson epipolar error threshold of $\tau_{\text{epi}} = 3$ and a segmentation-confidence threshold of $\tau_{\text{mask}} = 0.8$, both determined empirically.

All frames are still used for per-point tracking and scene-flow estimation, which improves trajectory accuracy with minimal additional computational cost.
For the DyCheck~\cite{gao2022dynamic} and NVIDIA (Gaussian Marbles)~\cite{stearns2024marbles} datasets, point trajectories are extracted using BootsTAPIR~\cite{doersch2024bootstap}.
However, for the original NVIDIA dataset~\cite{9156445}, following MoSca~\cite{11094310}, we use CoTracker~\cite{cotracker}, which provides more stable tracking under sparse temporal sampling than BootsTAPIR~\cite{doersch2024bootstap} (see Table~\ref{tab:tracker_comp}).

\begin{table}[h]
\begin{center}
\begin{tabular}{l|ccc}
\toprule
\textbf{Tracker} & \textbf{PSNR}$\uparrow$ & \textbf{SSIM}$\uparrow$ & \textbf{LPIPS}$\downarrow$ \\
\midrule
BootsTAPIR & 26.22 & 0.865 & 0.062 \\
CoTracker & \textbf{26.26} & \textbf{0.866} & \textbf{0.060} \\
\bottomrule
\end{tabular}
\end{center}
\caption{Quantitative evaluation of novel view synthesis results obtained with different tracking methods on the original NVIDIA dataset~\cite{9156445}.}
\label{tab:tracker_comp}
\end{table}

\subsection{Trajectories Encoding}
For trajectory encoding, we use a polynomial degree of $d_{\text{pol}} = 3$ and a Fourier degree of $d_{\text{Fourier}} = 32$ for long sequences in DyCheck~\cite{gao2022dynamic}.
For the NVIDIA (Gaussian Marbles) dataset~\cite{gao2022dynamic}, which contains shorter videos, a reduced degree of $d_{\text{Fourier}} = 24$ provides a more compact representation without sacrificing accuracy.
For the original NVIDIA dataset\cite{9156445}, which offers only 12 training frames, we further decrease the degree to $d_{\text{Fourier}} = 4$. This ensures that the number of temporal samples exceeds the number of unknown Poly-Fourier coefficients, which produces an overdetermined system that can be solved robustly using least squares.

\begin{figure*}[t]
\begin{center}
\includegraphics[width=0.97\textwidth]{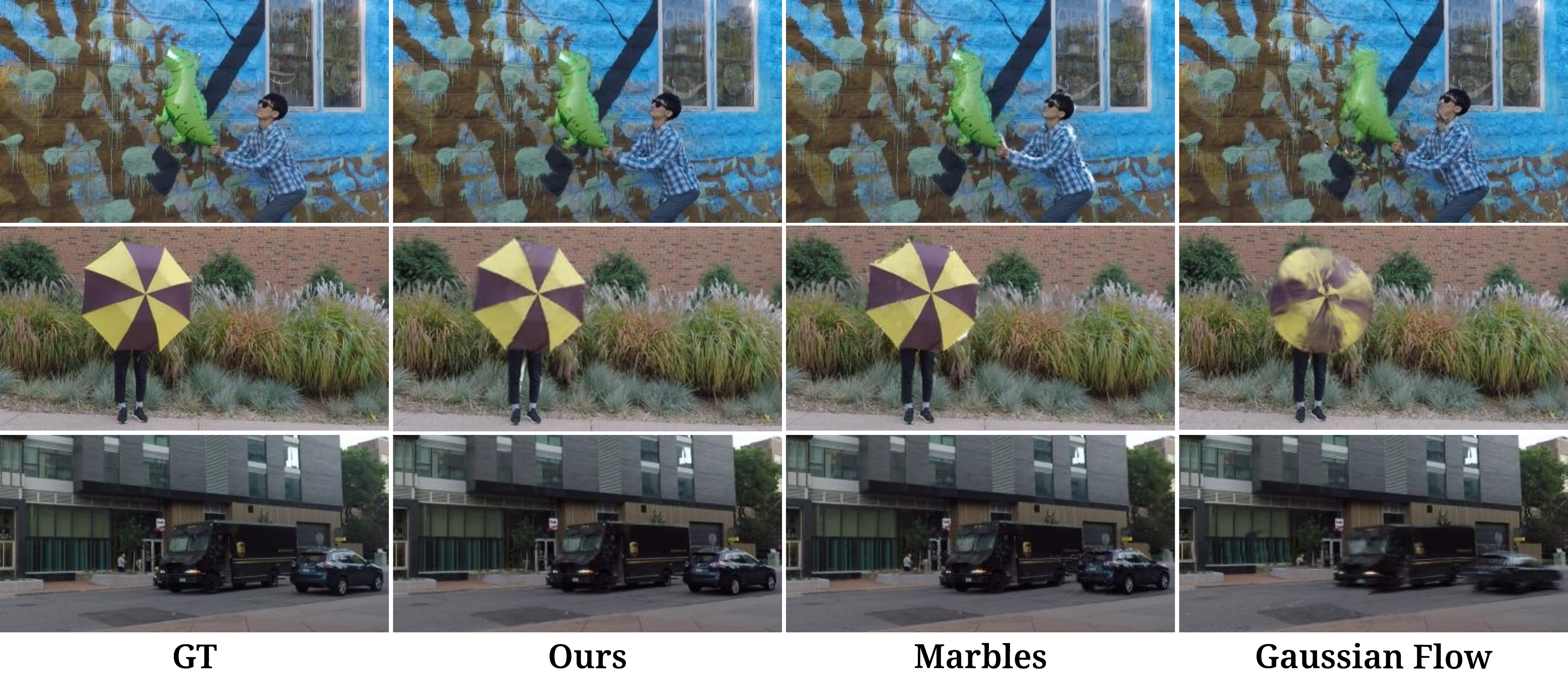}
\end{center}
    \caption{\textbf{Qualitative comparison on the NVIDIA (Gaussian Marbles) dataset~\cite{stearns2024marbles}.}
MOSAIC-GS produces smoother, more coherent object structures and cleaner contours, whereas Gaussian Marbles exhibits grainy artifacts and irregular, patchy boundaries in dynamic regions.}
\label{fig:nvidia_gm_comparison}
\end{figure*}

\subsection{Initialization of Static Regions}
We sample points from static regions every 20th frame, which provides broad scene coverage without excessive oversampling. For the original NVIDIA dataset~\cite{9156445}, which contains only 12 spatially distributed frames, we instead sample points from every second frame.

Our sampling follows a slightly modified version of the probability-based strategy proposed by Meuleman \etal~\cite{meuleman2025onthefly}.
For each pixel $(x, y)$ in frame $\mathbf{I}_t$, we compute the Laplacian-of-Gaussian magnitude $\text{LoG}(x,y)$. The sampling probability is then defined as

\begin{equation}
P(x,y) =
\begin{cases}
\text{LoG}(x,y), & \text{if } \mathbf{M}_t(x,y) = 0, \\[6pt]
0, & \text{if } \mathbf{M}_t(x,y) > 0,
\end{cases}
\end{equation}

where $\mathbf{M}_t$ denotes the combined segmentation mask for frame $t$, with the value $0$ indicating static regions.

\subsection{Photometric Optimization}
We perform photometric optimization for 20,000 iterations.  
The loss function combines photometric losses L1 and SSIM with $\lambda_{\text{ssim}}\ = 0.2$ and depth regularization loss with $\lambda_{\text{depth}}\ = 0.2$. Although our method supports spherical harmonics, in our experiments we set the degree to 0, using only RGB values for Gaussian colors. We found that this provides sufficient representation capacity on monocular benchmarks while remaining more compact than higher-degree spherical harmonics.

The model is trained using the Adam~\cite{adam} optimizer with separate learning rates for different Gaussian parameters. 
For deformation parameters, we use an initial learning rate of $5 \times 10^{-6}$ for the mean offsets, decaying exponentially to $5 \times 10^{-8}$, and $1 \times 10^{-6}$ for rotation quaternions, decaying to $1 \times 10^{-8}$. 
All other optimization settings follow the original Gaussian Splatting scheme~\cite{kerbl3Dgaussians}.

For the NVIDIA datasets~\cite{9156445, stearns2024marbles}, scenes are rescaled to the $[-1,1]$ cube. 
For DyCheck~\cite{gao2022dynamic}, we use the refined camera poses provided by MoSca~\cite{11094310} without additional scaling.  
All experiments were conducted on a single NVIDIA RTX 4090 GPU with 24 GB of memory.

\section{Additional Experiments}
In addition to the experiments presented in the main paper, we include a qualitative comparison of MOSAIC-GS with prior methods on the NVIDIA (Gaussian Marbles) dataset~\cite{stearns2024marbles}, as illustrated in Fig.~\ref{fig:nvidia_gm_comparison}.

We also provide a per-scene quantitative comparison across both variants of the NVIDIA datasets~\cite{9156445, stearns2024marbles}, summarized in Table~\ref{tab:nvidia_full_comparison}. Additionally, the $video\_results$ folder in our supplementary materials contains videos of MOSAIC-GS reconstruction results captured from the evaluation cameras for scenes from the NVIDIA (Gaussian Marbles)~\cite{stearns2024marbles} and DyCheck~\cite{gao2022dynamic} datasets.

\begin{table*}[t]
\begin{center}
\resizebox{\textwidth}{!}{ 
\begin{tabular}{l|ccccccc|c}
\toprule
\multicolumn{7}{l}{\textit{NVIDIA (origin)}\cite{9156445}} & \multicolumn{2}{r}{\textit{Reported: PSNR$\uparrow$ / LPIPS$\downarrow$}}\\
\rule{0pt}{1ex}   
 & Balloon1 & Balloon2 & Jumping & Playground & Skating & Truck & Umbrella & Mean \\
\midrule
Gaussian Flow\cite{10658506} & 20.98 / 0.199 & 23.15 / 0.152 & 23.13 / 0.153 & 18.42 / 0.170 & 26.57 / 0.099 & 25.92 / 0.075 & 22.60 / 0.189 & 22.97 / 0.148 \\
MoSca\cite{11094310} & \textbf{23.58} / \underline{0.10} & \textbf{27.80} / \underline{0.05} & \textbf{25.01} / \underline{0.09} & \textbf{24.25} / \underline{0.05} & \textbf{33.41} / \textbf{0.03} & \textbf{27.83} / \underline{0.08} & \underline{25.17} / \underline{0.09} & \textbf{26.72} / \underline{0.07} \\
Ours & \underline{23.21} / \textbf{0.090} & \underline{27.77} / \textbf{0.040} & \underline{24.44} / \textbf{0.085} & \underline{24.14} / \textbf{0.049} & \underline{31.73} / \underline{0.032} & \underline{27.24} / \textbf{0.060} & \textbf{25.27} / \textbf{0.066} & \underline{26.26} / \textbf{0.060} \\
\midrule
\noalign{\vspace{3pt}}
\multicolumn{7}{l}{\textit{NVIDIA (Gaussian Marbles)}\cite{stearns2024marbles}} & \multicolumn{2}{r}{\textit{Reported: PSNR$\uparrow$ / LPIPS$\downarrow$}}\\
\rule{0pt}{1ex}   
 & Balloon1 & Balloon2 & Jumping & Playground & Skating & Truck & Umbrella & Mean \\
\midrule
Gaussian Flow\cite{10658506} & 22.24 / 0.068 & 21.72 / 0.116 & 19.28 / 0.123 & 17.05 / 0.134 & 26.14 / 0.039 & 24.39 / 0.092 & 22.41 / 0.095 & 21.89 / 0.095 \\
Gaussian Marbles\cite{stearns2024marbles} & \underline{24.09} / \underline{0.041} & \textbf{23.84} / \underline{0.077} & \underline{20.20} / \textbf{0.100} & \underline{17.48} / \underline{0.127} & \textbf{27.83} / \textbf{0.030} & \underline{27.30} / \underline{0.049} & \textbf{25.04} / \textbf{0.059} & \underline{23.68} / \underline{0.069} \\
Ours & \textbf{24.44} / \textbf{0.036} & \underline{23.76} / \underline{0.077} & \textbf{20.70} / \underline{0.101} & \textbf{17.54} / \textbf{0.126} & \underline{27.69} / \underline{0.032} & \textbf{27.82} / \textbf{0.044} & \underline{24.59} / \underline{0.066} & \textbf{23.79} / \underline{0.069} \\
\bottomrule
\end{tabular}}
\end{center}
\caption{\textbf{Per-scene quantitative results on NVIDIA datasets.} MOSAIC-GS achieves the lowest LPIPS scores on the original NVIDIA dataset \cite{9156445} and establishes a new state-of-the-art on the NVIDIA (Gaussian Marbles) dataset \cite{stearns2024marbles}.}
\label{tab:nvidia_full_comparison}
\end{table*}

\begin{figure}[t]
\begin{center}
\includegraphics[width=\linewidth]{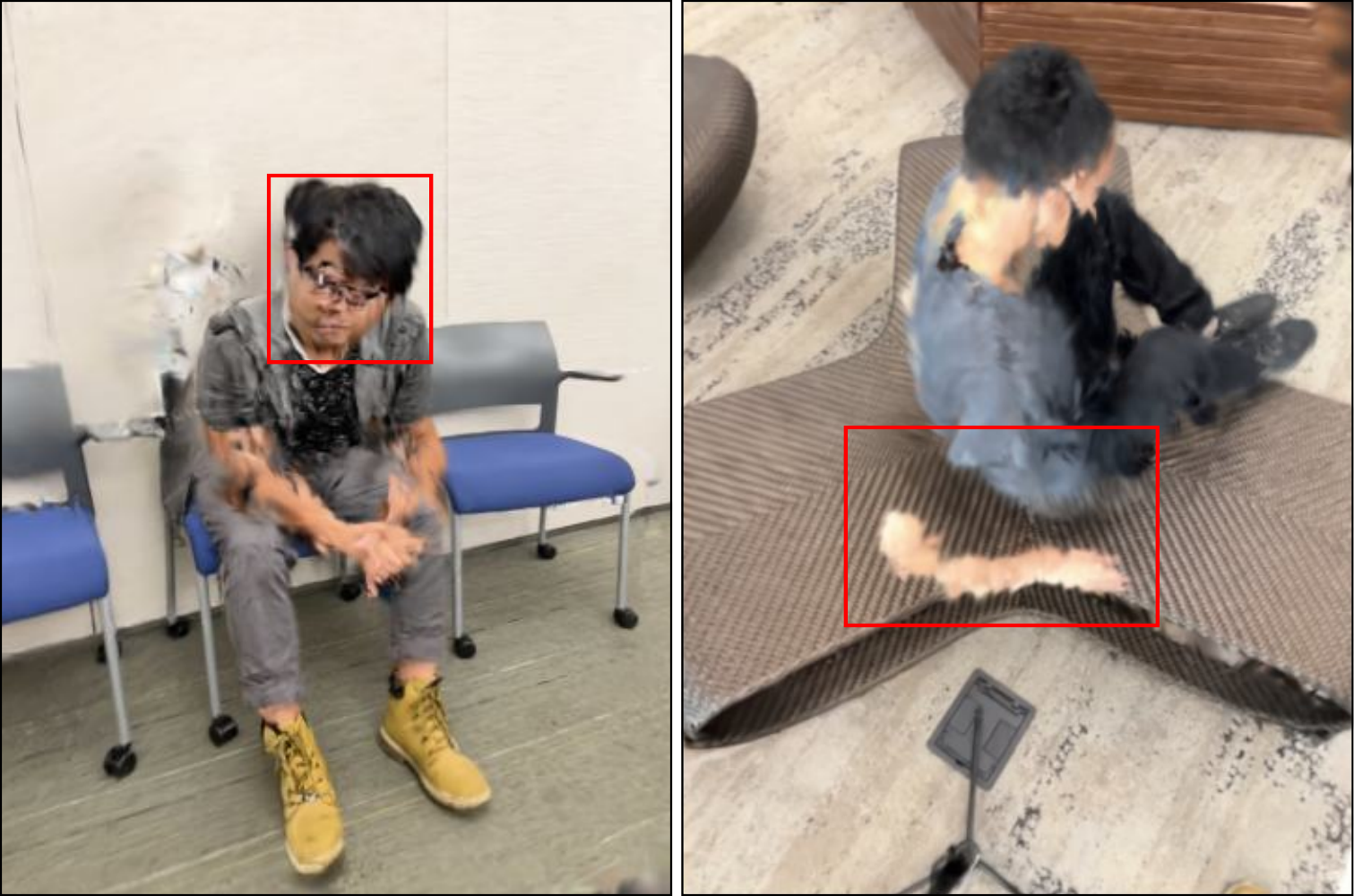}
\end{center}
    \caption{\textbf{Illustration of two key limitations of MOSAIC-GS.} 
\textbf{Left:} Insufficient scene flow refinement leads to missing fine facial details when the person turns away from the camera. 
\textbf{Right:} Splitting a single object into separate instances prevents the occluded hand from inheriting motion cues from the visible body, leading to incorrect motion estimation.}
\label{fig:limitations}
\end{figure}

\section{Limitations and Future Work}
As discussed in the main paper, MOSAIC-GS has two primary limitations: (1) reliance on accurate initialization, and (2) insufficient scene flow refinement in certain cases.
 
First, our method depends on the quality of initial scene flow estimation and segmentation masks. If dynamic regions are not correctly identified during initialization, photometric optimization may fail to capture their motion accurately, leading to reconstruction artifacts. 
Because the approach relies on external models for optical flow, segmentation, and point tracking, it may also inherit their limitations. However, ongoing advances in these areas are likely to directly benefit our method.

Second, when dynamic objects are completely invisible in certain training frames, the system may struggle to reconstruct fine details of their appearance or motion. The examples for this type of limitation are presented in Fig.~\ref{fig:limitations}.
This issue mainly stems from two factors: (1) limited precision of scene flow refinement during initialization, and (2) segmentation of a single object into multiple instances (e.g., human body parts segmented separately). 
The first issue could be mitigated by introducing rigidity constraints during photometric optimization, while the second opens an interesting direction for future work on leveraging correlations among dynamic instances to infer the motion of unobserved parts from related visible ones.

{\small
\bibliographystyle{ieee_fullname}
\bibliography{egbib}
}